# F$^3$T: A soft tactile unit with 3D force and temperature mathematical decoupling ability for robots

Xiong Yang[1], Hao Ren[2], Dong Guo[2], Zhengrong Ling[1], Tieshan Zhang[2], Gen Li[2], Yifeng Tang[2], Haoxiang Zhao[1], Jiale Wang[1], Hongyuan Chang[1], Jia Dong[1], Yajing Shen[1,3]

1.  Department of Electronic and Computer Engineering, The Hong Kong University of Science and Technology, Hong Kong, China
2.  Department of Biomedical Engineering, City University of Hong Kong, Hong Kong, China
3.  Center for Smart Manufacturing, The Hong Kong University of Science and Technology, Hong Kong, China

ABSTRACT:
The human skin exhibits remarkable capability to perceive contact forces and environmental temperatures, providing intricate information essential for nuanced manipulation. Despite recent advancements in soft tactile sensors, a significant challenge remains in accurately decoupling signals - specifically, separating force from directional orientation and temperature - resulting in fail to meet the advanced application requirements of robots. This research proposes a multi-layered soft sensor unit (F$^3$T) designed to achieve isolated measurements and mathematical decoupling of normal pressure, omnidirectional tangential forces, and temperature. We developed a circular coaxial magnetic film featuring a floating-mountain multi-layer capacitor, facilitating the physical decoupling of normal and tangential forces in all directions. Additionally, we incorporated an ion gel-based temperature sensing film atop the tactile sensor. This sensor is resilient to external pressure and deformation, enabling it to measure temperature and, crucially, eliminate capacitor errors induced by environmental temperature changes. This innovative design allows for the decoupled measurement of multiple signals, paving the way for advancements in higher-level robot motion control, autonomous decision-making, and task planning.

MAIN TEXT:
## INTRODUCTION
To equip robots with human like environmental perception and adaptive interaction abilities, tactile sensors with multi-signal sensing capabilities - particularly for force and temperature - are essential and have garnered significant attention recently [1-7]. To date, various types of soft tactile sensors have been proposed, which can be roughly categorized into electrical, ion gel, optical, and magnetic signal-based sensors [8-22] (Table 1, Supporting Materials Section S1). These sensors have demonstrated considerable application potential in force measurement, with some exhibiting skin-comparable sensitivity [17,22]. However, it is essential to recognize that force is a vector comprising both magnitude and direction components, and accurately determining these components is crucial for estimating operational states and taking appropriate actions. Nevertheless, due to signal coupling and interference in existing soft tactile sensors, accurately decoding normal force and omnidirectional tangential force remains a significant challenge, particularly when considering temperature effects.

To address the issue of signal decoupling, scientists have made several efforts over the years, focusing on materials, structures, and principles. For instance, stretchable sensors based on carbon nanotubes and graphene oxide have been developed to decouple the 1D tangential force and normal pressure [11]. Sensor arrays utilizing 3D field-coupled thin-film transistors enable simultaneous measurement of the spatial distribution of normal pressure and temperature [12]. Additionally, sensing receptors based on ion relaxation dynamics can distinguish thermal information from mechanical information without signal interference [17]. Recently, Liu et al. introduced a 3D force decoupling electronic skin (3DAE-Skin) using a heterogeneous encapsulation strategy, which incorporates pressure and strain sensors oriented in different directions to measure multidimensional forces [14]. They demonstrated its applications in robotic manipulation and assessing food freshness. However, its piezoresistive-based 3D array design leads to reduced accuracy and stability, higher structural complexity, increased production costs, and an inability to decouple temperature effects on measurements.

Previously, our group proposed a Halbach-based magnetic film and demonstrated its theoretical capability to decouple normal force and orientation-fixed shear force [22]. However, achieving decoupled normal and omnidirectional tangential force sensing continues to be a challenge. Moreover, the electrical components of tactile sensors are typically sensitive to temperature changes, while robots frequently manipulate objects with varying temperatures [23-29]. The development of a highly integrated, compact, low-cost soft tactile sensing unit with high-precision three-dimensional force and temperature decoupling capabilities is imperative for achieving adaptive, precise, and intelligent robotic operations.

In this work, we propose a skin-inspired, multi-material four-layer sensory unit capable of mathematically decoupling 3D force and temperature, referred to as the $F^3T$ unit (Fig. 1, Supporting Materials Fig. S1 and Fig. S2). It dynamically measures normal force, tangential force, and temperature from operated targets using electromigration, magnetic induction, and ion translocation, respectively. We demonstrate its ability to decouple static signals and recognize dynamic operation in intelligent robotic manipulation and human-robot cooperation. This advancement opens up significant possibilities for enhanced environmental perception, motion control, autonomous decision-making, and task planning for robots [30-35].

**Table S1. Comparison of state-of-the-art flexible sensors in force and temperature measurement decoupling.**

| Sensing principles | Sensors | 2D tangential force decouple | 1D tangential-normal force decouple | 2D tangential-normal force decouple | 3D force - temperature decouple | Ref |
|---|---|---|---|---|---|---|
| Electrical signal based | CNTs/GO e-skin | No | No | No | No | 8 |
| | FP-TS | No | No | No | No | 9 |
| | Planar capacitor arrays | Yes | Yes | Yes | No | 10 |
| | PET substrate e-skin | No | No | No | No | 11 |
| | FCTFT sensor | No | No | No | No | 12 |
| | 3D capacitor e-skin | Yes | No | No | No | 13 |
| | 3DAE-Skin | Yes | Yes | Yes | No | 14 |
| | Zwitterionic Skins | No | No | No | No | 15 |
| | Iontronic skins | No | No | No | No | 16 |
| | AMI receptor | Yes | No | No | No | 17 |
| Optical based | TacTip | Yes | Yes | Yes | No | 18 |
| | Optoelectronic TF unit | Yes | Yes | Yes | No | 19 |
| | GelSight | Yes | Yes | Yes | No | 20 |
| Magnetic field based | Traditional magnetic field based | Yes | No | No | No | 21 |
| | Soft M-skin | No | Yes | No | No | 22 |
| Compound | F$^3$T unit (This work) | **Yes** | **Yes** | **Yes** | **Yes** | This work |

## RESULTS

### Skin like multi-layer design of the F³T sensory unit

The skin is the largest organ in the human body, featuring a multi-layer structure composed of the epidermis, dermis, and hypodermis, each containing specialized receptors with distinct functions (Fig. 1a). Mechanoreceptors, categorized into Pacinian corpuscles, Meissner corpuscles, Krause endings, Ruffini endings and so on, enable the sensation of 3D force, while thermoreceptors detect temperature. This unique architecture allows human skin to effectively decouple different signals, facilitating accurate recognition of both the magnitude and direction of force, as well as the temperature, without interference from one another even in complex environmental conditions.

To achieve skin-like decoupling abilities, we designed a tactile sensing unit (F³T) with multiple layers (Fig 1b). The topmost layer (1 mm thick) is created by pouring an ion gel mixture (LPEGDA: LHOMPP: PBS Buffer = 5:5:1) onto a porous mesh fabric. After curing under UV light, the ion gel film is demolded and encapsulated with cyanoacrylate to improve reliability during contact (Fig. 1d). The second layer consists of a thin magnetic film with a circular Halbach pattern (0.5 mm thick). To create this layer, we first fabricate a rectangular magnetic film with a standard Halbach array magnetization by mixing polydimethylsiloxane (PDMS) and neodymium (NdFeB) magnetic powders. We then use a laser to cut the film into small isosceles triangles with a vertex angle of 22.5 degrees, which are assembled into a circular Halbach magnetic film (Fig. 1e). The third layer (5 mm thick) is a floating capacitor composed of four circular electrodes, with a gap of 1 mm between each. The diameters of the electrodes decrease from bottom to up, measuring 9 mm, 7.5 mm, 6 mm, and 4.5 mm, respectively. A silicone elastomer containing phosphate ions serves as the medium between the electrodes, molded to fit precisely (Fig. 1f). This design ensures that the capacitor is sensitive only to the applied normal force, effectively eliminating the influence of tangential force.

Consequently, the F³T unit is fabricated into a compact cylindrical structure with a radius of 6 mm and a height of 8 mm. It features an ionic gel-based top layer for temperature sensing via ion translocation, a circular Halbach magnetic film-based middle layer for tangential force sensing through magnetic induction, and a floating capacitor-based bottom layer for normal force sensing via electromigration. The inner silicone elastomer resembles the hypodermis of human skin, facilitating connections between components and buffering impact forces. Additionally, a rigid printed circuit board (PCB) with a 3D Hall sensor serves as the backbone, providing fixation and support for the F³T sensing unit, analogous to the bone structure of a finger. Electrical signals from the temperature-sensitive ion gel layer and the floating capacitor are transmitted to the bottom PCB via two pairs of thin wires. More details about the architecture can be found in the Supporting Materials Section S2.

### Decoupling principle for the temperature, normal force, and all-directional

**tangent force**

Unlike existing multi-signal sensing units that struggle to accurately measure and decouple 3D force and temperature, our multi-layer design allows for mathematical decoupling these signals. As illustrated in Fig. 2a, when the F$^3$T sensor contacts a target object, a composite signal denoted as $f(T, Fz, Fx, Fy)$, which includes temperature, normal force and tangential force, is transmitted to the sensor. The topmost ion gel layer directly contacts the object to gather temperature information. When the temperature changes, the distance between polymer chains within the gel also changes, leading to the conversion between non-free ions and free ions, resulting in a change in resistance (Supporting Materials Section S3, Fig. 2b). Consequently, the temperature $T$ of the target object can be effectively decoupled from the force $F$ and accurately measured.

The floating capacitor comprises four intersecting disc electrodes (2 positive and 2 negative) with varying radii, made of soft elastomer as the medium. The total measured capacitance is given by $C = C_{12} + C_{14} + C_{23} + C_{34} = \varepsilon r(3S_1 + 2S_2 + 2S_3)/8\pi k d$ (Supporting Materials Section S4 and Supporting Materials Fig. S3). This capacitance is insensitive to tangential force components since the effective capacitor area $S$ and electrode distance $d$ do not change under tangential load. Noted that both the electrodes distance $d$ and relative permittivity $\varepsilon r$ are correlated to the temperature $T$ due to thermal expansion (cte: $d = f(Fn, cte)$) and ionization (it: $\varepsilon r = f(it)$) properties of the material. Since these factors have opposing effects on capacitance $C$, we can design the cte/it ratio of the material to achieve temperature-sensitive adaptive compensation by regulating the ion content in the soft elastomer (Fig. 2c and Supporting Materials Section S4). Consequently, change in capacitance $C$ are solely related to variations in electrode distance $d$ caused by normal force, allowing for accurate measurement of normal force.

The magnetic flux distribution on the Hall sensor is generated by the magnetic film and is correlated to the deformation of the soft elastic layer caused by either tangential or normal force. Additionally, the measurement accuracy of the Hall sensor's magnetic field is affected by temperature changes. Thus, the initially measured magnetic field can be represented as $M = f(T, Fz, Fx, Fy)$ (Fig. 2d). For the circular coaxial Halbach magnetic film, its tangential magnetic field distribution $M_r$ can be expressed as:

$$M_r = M_0(T, F_z)\sin\left(k\sqrt{\Delta x^2 + \Delta y^2}\right) \quad (1)$$

where $M_0(T, F_z)$ is the tangential magnetic field strength amplitude as a function of temperature $T$ and normal force $Fz$. In the previous step, the temperature $T_0$ and normal force F$_{z0}$ were obtained independently through ion gel and floating capacitor. These values can then be substituted into equation (1) to obtain the distribution of the tangential magnetic field $M_r$ with displacement $\Delta x$ and $\Delta y$ under specific conditions of temperature $T_0$ and normal force $F_{z0}$. Since the displacement in the $x$ and $y$ directions can be directly converted into tangential force $F_x$ and $F_y$ using shear strain $\gamma$ and shear modulus $G$, the relationship between tangential force $F_x$, $F_y$ and the tangential magnetic field $M_r$ can be uniquely determined:

$$F_x = S \cdot (G \cdot \gamma_x) = S \cdot G \cdot \frac{f(M_r)}{h} \cos\theta_r \quad (2)$$

$$F_y = S \cdot (G \cdot \gamma_y) = S \cdot G \cdot \frac{f(M_r)}{h} \sin\theta_r \quad (3)$$

Furthermore, the calculation of normal force and tangential force is closely related to the material properties of the elastomer, allowing for adjustments in the measurement range and accuracy as needed. Generally, the smaller the elastic modulus $E$ and shear modulus $G$ of the material result in a reduced sensor range and increased measurement accuracy. For this study's characterization, the elastomer used in the sensor was dragon skin (dragon skin 30, Smooth-On, Inc.) doped with phosphate ions.

### Characterization of F³T in decoupling multiple tactile signal

To calibrate the relationship between contact temperature $T$ and the current $I$ within the gel, we developed an equivalent voltage divider circuit on the bottom PCB substrate. As shown in Fig. 3a, the current $I$ exponentially increases with the increase of temperature $T$, modeled by a polynomial equation (Supporting Materials Section S5) with an error of less than $\pm 0.5$ °C within the range of 20 - 80 °C. We measured the temperature $T$ under different normal loading (from 0 to 7 N) and tangential loading (from 0 to 2 N) conditions. We then calculated the ratio of the measured temperature under force loading conditions to the temperature at free load condition. As shown in Fig. 3b, the ratio remained close to 1 with an error of less than 1% while environmental temperature changed from 20 to 80 °C. This verifies that the effect of external force $F$ on temperature measurement is negligible, consistent with theoretical prediction.

The normal force $F_z$ is measured based on the value of the floating capacitor $C$. Here, we convert the capacitance signal measurement into a frequency signal $f$ using an NE555 timer IC circuit to increase accuracy (Fig. 3c, Supporting Materials Section S4). The calibration result indicates that $f$ is approximately linear to the normal displacement $\Delta H$ with a slope of -900 Hz/mm (Fig. 3d), while it has two linear segments corresponding to $F_z$. This behavior reflects changes in the apparent stiffness due to different deformation regimes: initially elastic at small deformations and more complex responses at larger deformations, due to geometric effects, contact conditions, or changes in internal structure (Fig. 3e). By using piecewise linear functions with slopes of -1/600 N/Hz and -1/200 N/Hz to model the relationship between the normal force $F_z$ and the output frequency $f$ (Supporting Materials Section S6), we effectively decouple the response into distinct linear regions. This approach achieves a measurement accuracy of 0.01 N in the range from 0 to 1 N and 0.03 N for values greater than 1 N. Furthermore, experimental results suggest that the capacitance $C$ is independent of environment temperature $T$ and tangential force (Fig. 3f). These characterization results verify the effectiveness of the mounting capacitance design for the decoupled normal force $F_z$ measurement.

To evaluate the accuracy of the tangential force $F_\tau$, we first characterized the tangential magnetic field distribution $M_r$. The results show that the strength of $M_r$ varies sinusoidally with tangential displacement $\Delta R$ at a fixed normal displacement $\Delta H$ (Fig. 3g, 3h, and Supporting Materials Fig. S4), which is a result of the circular Halbach magnetization pattern we used. Within the defined tangential force measurement range ($\Delta R \leq 2.5$ mm), the intensity of the magnetic field increases monotonically with $\Delta R$ over the ascending part of the sine wave. This behavior indicates a unique correspondence between $\Delta R$ and $M_r$ when $\Delta H$ is known, allowing us to establish a mathematical model to calculate the tangential force $F_\tau$ (Fig. 3i, Supporting Materials Section S7 and Supporting Materials Fig. S5).

When a normal force $F_z$=7.2 N is applied (i.e. $\Delta H = 2$ mm), the calibration results show that the tangential magnetic field increases with tangential force initially, with an average slope of $K1 = 1/10531$ N/uT. When tangential displacement $\Delta R > 2$ mm (i.e. tangential force greater than 1 N), the average slope changes to $K2 = 1/377$ N/uT due to the sinusoidal distribution characteristics of the tangential magnetic field. Notably, leveraging this characteristic, the magnetic field changes more significantly in the initial phase, providing the sensor with higher sensitivity (0.01 N) to measure $F_\tau$ even when the load is small (0 - 1N). The orientation of the tangential force $\theta$ is determined by the signs of the tangential magnetic fields in the x and y directions, specifically using the arctan of the ratio $B_y/B_x$. Calibration results indicate that the $\theta$ is independent of the normal force $F_z$ (Fig. 3j) and the magnitude of $F_\tau$ (Fig. 3k), which aligns well with actual situation, so $\theta$ can be expressed with piecewise linear functions (Supporting Materials Section S8 and Supporting Materials Fig. S6). Furthermore, the respective component strengths of the tangential force in the x and y directions can be calculated by $F_x = F_\tau \cdot \cos\theta$ and $F_y = F_\tau \cdot \sin\theta$ respectively.

The influence of temperature $T$ on magnetic field measurement $M_x$ is calibrated and shown in Fig. 3l, which exhibits a linear relationship. Thus, the influence of $T$ on the magnetic field can be mathematically normalized to achieve compensation and correction during the measurement of tangential force. Consequently, we can achieve the mathematical decoupling of temperature $T$, normal force $F_z$, tangential force $F_x$, and tangential force $F_y$ by the F³T sensor.

**Performance evaluation of the F3T sensor under static and dynamic conditions**
Effective human-robot interaction requires not only precise, interference-free measurement of tactile signals under static conditions but also the ability to dynamically decouple these signals to accurately recognize human actions and intentions. While existing soft tactile sensors can be highly sensitive, they often struggle with decoupling forces from different directions due to signal interference. This limitation hinders the recognition of human actions, especially in subtle interactions, such as gentle taps, sliding motions, or object handovers. To demonstrate the capabilities of the F³T sensor, we evaluated its performance in multi-signal measurement and decoupling under both static and dynamic conditions.

To begin with, we first evaluated its static performance. As shown in Fig. 4a, a 100 g weight was placed on the sensor as a fixed normal loading at $t$=50 s. Subsequently, tangential forces were applied sequentially along the x-axis at $t$=100 s and along the y-axis at $t$=150 s. During this process, the ambient temperature progressively increased from 20 °C to 80 °C. For benchmarking purpose, we also conducted the same experiment using two widely used commercial sensors: the strain-force sensors (VISTE VSZ020) with an accuracy 0.01 N and the thermocouple temperature sensors (TEKTRONIX 2110-220) with an accuracy 0.25 °C.

As depicted in Fig. 4a, the temperature readings from the $F^3T$ sensor closely align with those from the standard thermocouple, demonstrating the $F^3T$'s robustness and stability (Average error less than $\pm 0.8$ °C, Supporting Materials Fig. S7, Fig. S8 and Fig. S9), even when subjected to variations in external loads. Regarding force measurement, the $F^3T$ maintained consistent readings as the temperature changed from 20°C to 80°C, with an average error of 0.03 N (within a 3% error margin). In contrast, traditional strain-force sensor exhibits significant thermal drift, with errors reaching up to 13.2% at 80°C. These results confirm that the $F^3T$ unit can accurately detect the magnitude and direction of 3D forces while simultaneously providing precise, decoupled measurements of both force and temperature in a static condition.

To evaluate the performance of the $F^3T$ sensor under dynamic conditions, we integrated it into a robotic gripper to hold a plastic block. We then use our hands to apply force with arbitrary magnitudes and specific directions, interacting with the robot by pulling up, down, left, right and pressing (Fig. 4b). During this process, the robot dynamically gained tactile information from the $F^3T$ sensor, and then accurately identified human movement direction and intention. Furthermore, by integrating tactile information as control feedback, grasper can achieve adaptive response to disturbances and various interactive actions from environment.

As shown in Fig. 4c, in the initial stationary phase (t=0-3 s), the gripper holds the target with a normal force of $F_z$ = 5.52 N and a y direction tangential force of $Fy$ = 0.80 N caused by the object's gravity ($Fx$ = 0 N). During this stage, the static grip force ($F_f=uF_z >$ $F_t=\sqrt{(F_x)^2+(F_y)^2}$) ensures the gripper maintains a stable hold on the object. In the jamming phase (3-7 s), a dynamic external pulling force with arbitrary direction is applied, increasing the tangential loads, which may exceed the static grip force $F_f$. To maintain stability and prevent unintended slipping, the gripper automatically increases the normal gripping force $F_z$ in response to the changes of 3D tangential force $F_t$. In this demonstration, the application of external interference makes the tangential load in y direction changes in range of 0.08 N to 1.56 N, while makes the tangential load in x direction changes in range of -0.62 N to 1.24 N. As a result, the normal gripping force of the gripper changes in range of 2.64 N to 8.28 N accordingly. This adjustment ensures that the static grip force from the gripper always exceeds the tangential load, allowing the grasper to maintain a stable grip on the target object. During the recovery phase (7-10 s), with the withdrawal of external interference, the tangential load is only in the x-direction and is the same as the initial value of 0.80 N. Likewise, the normal gripping force of the grasper also decreases, returning to the initial level of 5.52 N.

This adaptive grasping ability, not only allows the gripper to adapt to external disturbances, maintaining manipulation stability, but also allows the stable grasping with the smallest force, helps reduce power consumption and prevents excessive gripping force that could potentially damage the target object. These demonstrations highlight the $F^3T$ sensor's superior capability to handle diverse and dynamic 3D tactile interactions, which distinguishes it from other sensors that are typically limited to simpler scenarios like direct pressing.

**Demonstration of $F^3T$ for automated chemical reaction procedure**
Automatic laboratories play a crucial role in enhancing efficiency, accuracy, and productivity in both research and industry. They streamline processes, reduce human error, and ensure consistent experimental conditions, leading to faster results and improved reproducibility. However, one significant challenge in realizing automatic laboratories is the effective handling and manipulation of delicate and varied samples, which demands a high level of dexterity and precision. The soft tactile sensing unit with multi-signal decoupling capabilities with the potential to enhance robotic manipulation in automated laboratories by enabling precise 3D force measurements and temperature sensing, crucial for handling sensitive chemical tasks and maintaining optimal reaction conditions.

Polyvinyl Alcohol (PVA) is commonly used in laboratories to create hydrogels, adhesives, and coatings. However, high-degree hydrolyzed PVA is nearly insoluble in water at room temperature, typically requiring a combination of heating and stirring for complete dissolution. Yet, if the water temperature rises too rapidly or exceeds 90°C, there is a significant risk of agglomeration. To ensure consistency in experiments and maintain the performance of PVA-based materials, careful preparation is essential. This process involves real-time temperature monitoring, staged heating, and controlled shaking to distribute heat evenly, which is even challenge for novice technicians. To showcase the capabilities of $F^3T$ in an automatic laboratory setting, we demonstrate the automatic preparation of an 87% alcoholysis degree PVA solution using a robot arm equipped our sensor (Fig.5a).

As illustrated in Fig. 5b, 5c and Supporting Materials Video S1, we integrated the $F^3T$ tactile sensing unit into the robotic gripper. The gripper first approaches the beaker containing PVA particles and deionized water based on visual feedback within 6 s. Between 7 and 9 s, the gripper's posture and the distribution of tangential forces exerted on the beaker are detected and relayed in real-time. By analyzing the tangential force data, the system detects any imbalance and adjusts the gripper's position to ensure the beaker remains level.

Next, the beaker is transferred to a spirit lamp for heating, with the $F^3T$ sensor continuously monitoring the solution's temperature to reach the optimal 60-75°C (10 - 219 s). Unlike traditional thermometers, the $F^3T$ sensor seamlessly integrates with the robotic system, allowing for simultaneous temperature sensing and manipulation without manual intervention or separate equipment. This integration is crucial for maintaining a controlled heating process, preventing rapid temperature increases that could lead to agglomeration.

After reaching the target temperature, the robotic system removes the beaker from the heat source and initiates a controlled shaking operation at a frequency of 1 Hz (220 - 459 s). Throughout this phase, the $F^3T$ sensor monitors the beaker's orientation and the tangential forces acting on it, ensuring stable handling and preventing any unintended slippage or spillage.

This real-time feedback loop, enabled by the $F^3T$, is essential for maintaining precise control over the solution's agitation, which is critical for preventing PVA particle agglomeration. As the temperature gradually decreases from 75°C to around 60°C, the system pauses shaking and resumes heating to maintain the optimal dissolution environment. After approximately 15 mins of alternating heating and shaking cycles, the PVA particles fully dissolve without any agglomeration. This achievement underscores the exceptional flexibility of the $F^3T$ system, offering an adaptive task-solving strategy that is unmatched by standard chemical equipment.

**Demonstration of $F^3T$ for effective human-robot cooperation**
Beyond its role as a laboratory assistant, the robotic gripper integrated with the $F^3T$ soft tactile sensing unit also opens new possibilities for complex operations and more natural human-robot interactions in service scenarios. To illustrate its potential, we use the example of a "tea delivery" task—a simple and common activity for humans but a significant challenge for robots due to multiple dynamic interactions, such as varying liquid levels, cup orientation, temperature changes, and most importantly, sensing human intent for handover (Fig. 6 and Supporting Materials Video S2).

Initially, the robotic gripper, guided by visual positioning, successfully grasps a glass containing 20 ml of liquid at room temperature (18 °C) with a initial grasp force of 7.7 N (0–12 s). The robot then transports the glass to the water dispensing station (13–34 s) in preparation for pouring hot water. The pouring process is carefully divided into three stages: high-flow (35–45 s), low-flow (46–63 s), and micro-flow (64–113 s).

During the high-flow stage, the rapid pouring of hot water causes significant changes in tangential force (from 1.4 N to 2.9 N) in the direction of gravity and a quick rise in temperature from 18.5 °C to 25.7 °C in 10 s. As the process shifts to the low-flow stage, the hot water flow rate decreases, and the pouring includes brief pauses, resulting in slower changes—1.3 N in tangential force and a 14 °C rise over 17 s. Once the tea reaches 40°C, the process transitions to the micro-flow stage, where the hot water drips slowly into the cup. Here, changes in tangential force and temperature are minimal, allowing for precise temperature control and thermal equilibrium within the tea. When the temperature reaches 55°C, the pouring stops, and the cup is ready for delivery to the user (114–136 s). During the entire hot water pouring process, in order to maintain a stable grip, the gripping force Fn of the gripper gradually increased from the initial 7.7 N to 12.5 N.

Unlike traditional robots that rely on placing the cup on a flat surface for a non-contact handover, the robotic system equipped with the $F^3T$ sensor can detect subtle shifts in force and accurately interpret human intent. At 137–138 s, a rapid decrease in tangential force in the direction of gravity signals that the human hand has securely grasped the cup handle, enabling a smooth, natural handover.

The soft tactile sensing unit significantly enhances the robot's ability to recognize and adapt to human actions in real-time, far beyond the capabilities of traditional visual recognition, which can suffer from delays and occlusions. This breakthrough in integrating advanced tactile sensing technology into robotic systems paves the way for more sophisticated, intuitive, and human-like interactions, setting a new standard for the future of robotics in both industrial and service applications.

**DISSCUSSION**
In this work, we introduced an innovative soft tactile sensing unit, the $F^3T$, designed to enhance robotic manipulation by providing precise, decoupled measurements of normal and tangential forces as well as temperature. Inspired by the human skin's ability to perceive multiple stimuli, the $F^3T$ sensor features a multi-layered structure that integrates a circular coaxial magnetic film and a floating-mountain multi-layer capacitor for effective decoupling of force signals across all directions. Additionally, an ion gel-based temperature-sensing layer was incorporated to accurately monitor thermal changes without interference from external pressures or deformations. This unique design allows for accurate detection and mathematical decoupling of multiple signals, significantly advancing the robot's capabilities in real-time environmental sensing, complex manipulation tasks, and dynamic human-robot interactions. Through various experiments, including tasks like solution preparing and tea delivery, the $F^3T$ sensor demonstrated its ability to perform complex, delicate operations by recognizing human intentions and adapting its response accordingly. This technology represents a major step forward in developing more intuitive and adaptable robotic systems, facilitating a broader range of applications from laboratory automation to service scenarios, ultimately contributing to the advancement of robotics towards more human-like perception and interaction capabilities.

**Acknowledgments**

This work was support by Hong Kong RGC General Research Fund (16203923).

**Compliance with ethics guidelines**

Xiong Yang, Hao Ren, Dong Guo, Zhengrong Ling, Tieshan Zhang, Gen Li, Yifeng Tang, Haoxiang Zhao, Jiale Wang, Hong Yuan, Jia Dong, Yajing Shen declare that they have no conflict of interest or financial conflicts to disclose.

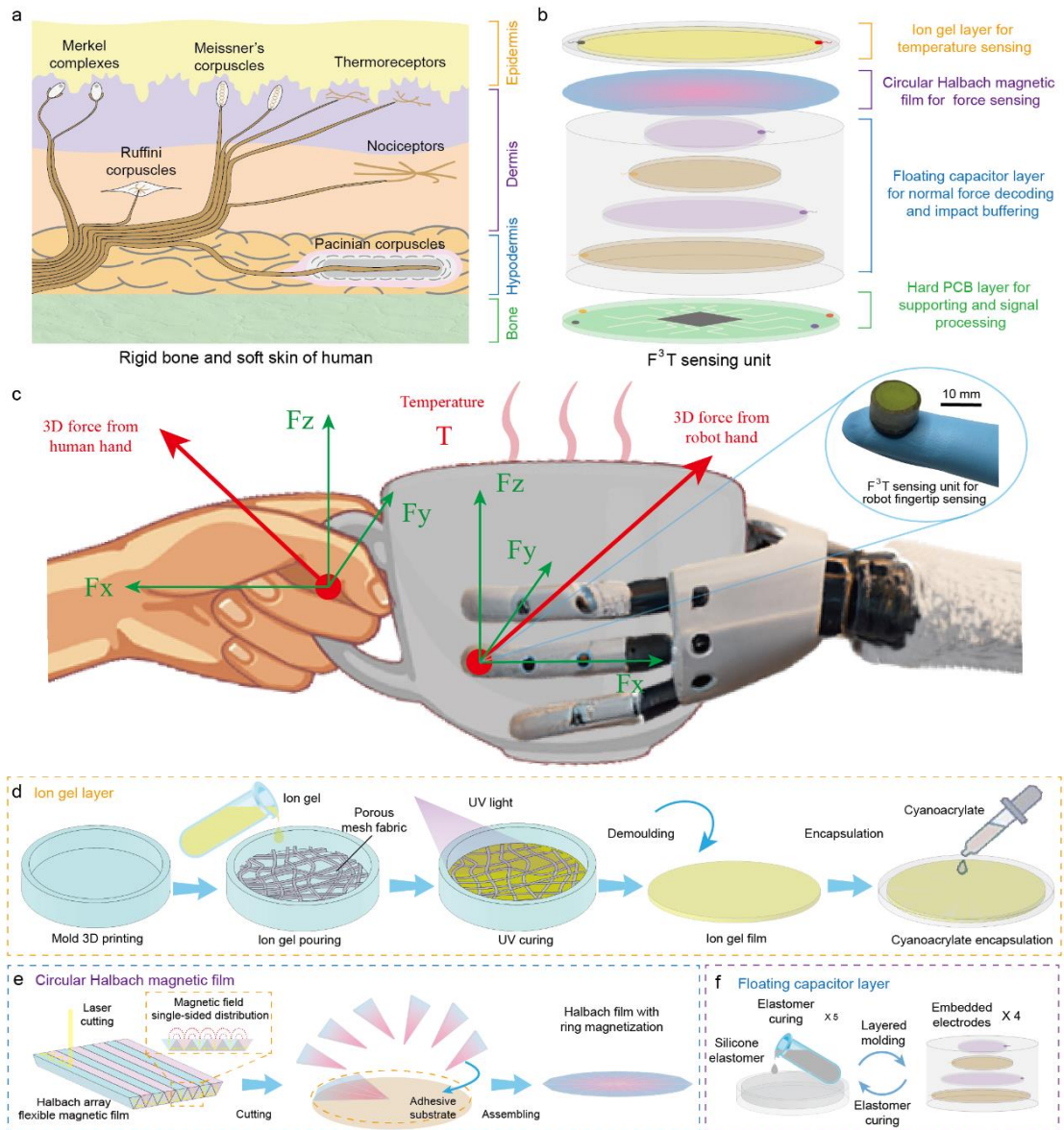

Figure 1. Overview of the sensing unit with decoupled force and temperature sensing ability. a. The multi-layered structure of human skin and various types of sensory receptors. b. The proposed multi-layered sensor unit structure inspired by the structure of human skin, capable of decoupled measurements of normal force, tangential force, and temperature. c. Sensing unit with multi-signal decoupling capabilities enable more intelligent and natural human-machine interaction. d. Fabrication of the temperature sensing layer that is insensitive to external forces. e. Assembly process of coaxial Halbach magnetic films for omnidirectional tangential force measurement. f. The manufacture of floating-mountain structure capacitance which is insensitive to tangential forces for normal force measurement.

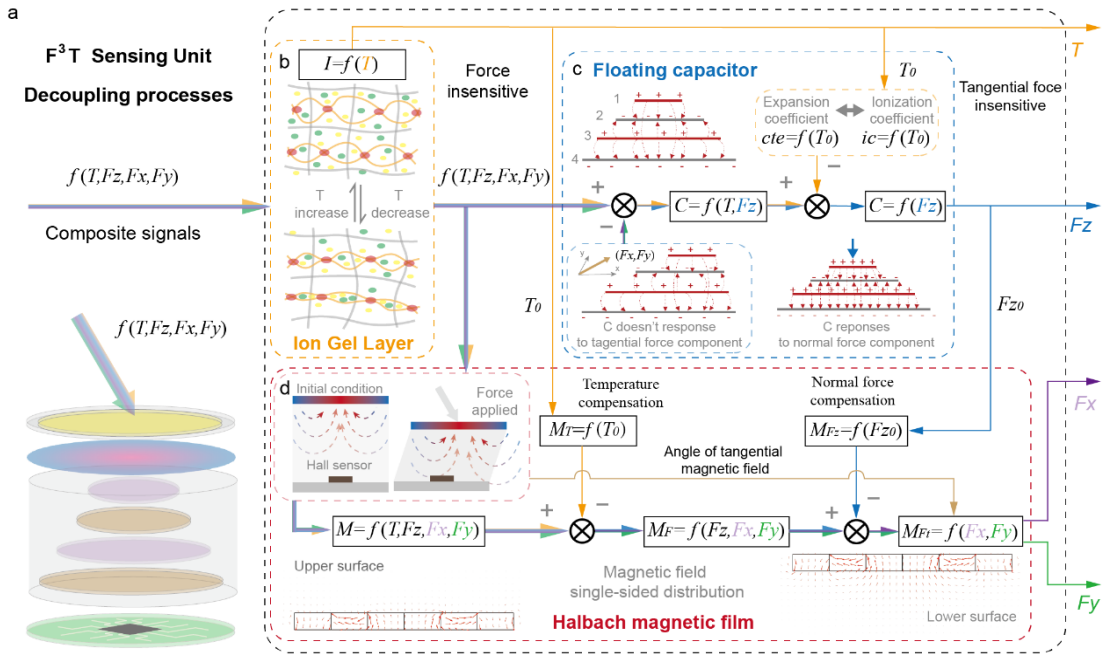

Figure 2. Signal decoupling process and principle of the F3T sensing unit. a. The signals applied to the sensor from the external environment include temperature and three-dimensional force information. b. The ion gel temperature sensing layer, which is sensitive to temperature but insensitive to force, first decouples and measures the temperature information independently. c. The floating-mountain capacitance removes tangential force interference through structural design and eliminates temperature interference through dielectric material regulation, ultimately achieving independent decoupled measurement of the normal force. d. Using the independently decoupled measurements of temperature and normal force, the corresponding relationship between the magnetic field distribution of the coaxial Halbach magnetic film and the omnidirectional tangential force can be further obtained.

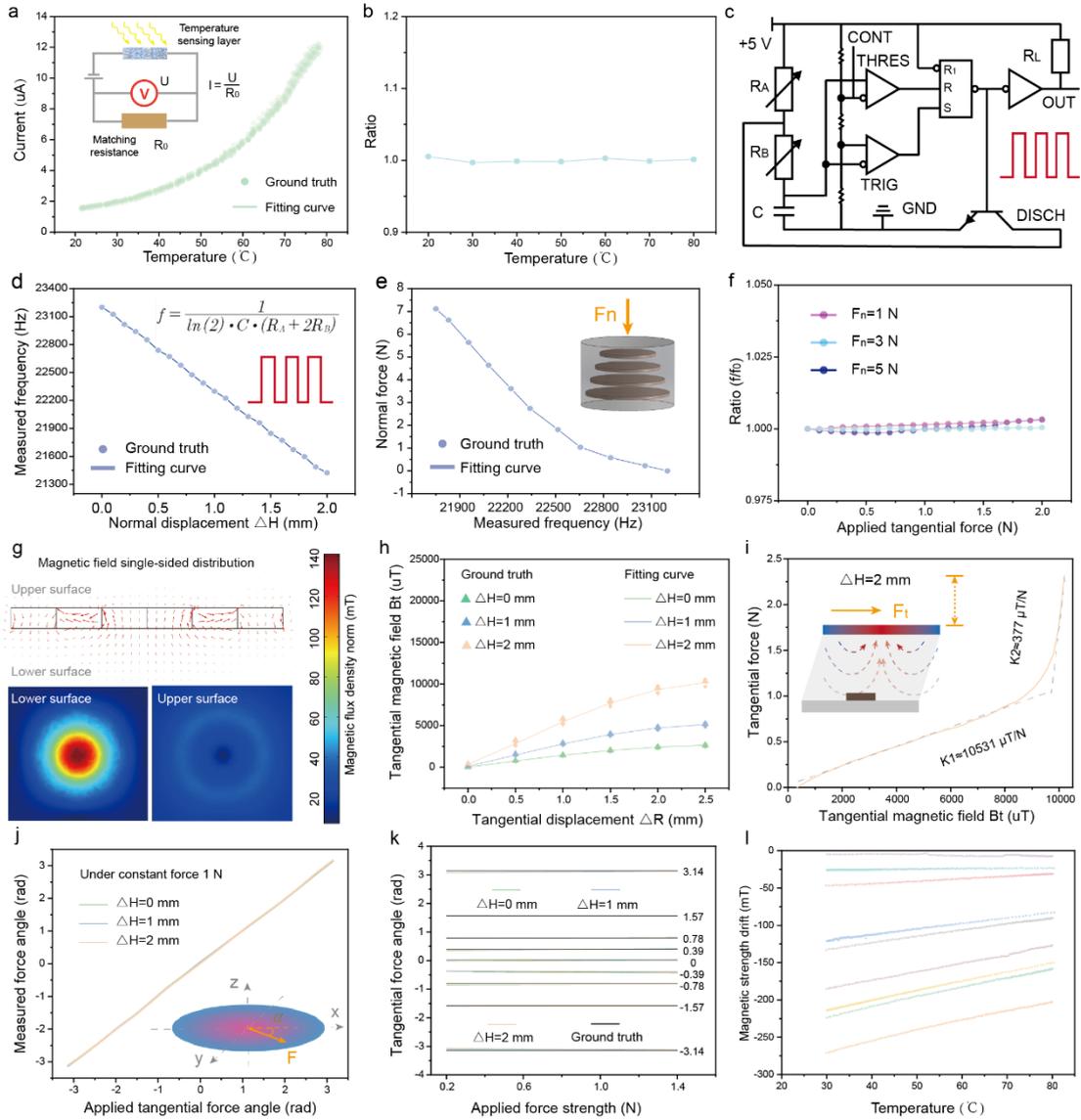

Figure 3. Characterization of the F³T sensing unit. a. Current variation of the ion gel-based temperature sensing layer at different temperatures. b. The ion gel-based temperature sensing layer shows insensitivity to force under different external load conditions. c. The NE555 timer IC circuit converts the capacitance signal measurement into a frequency signal. d. Linear relationship between normal displacement and the value of floating-mountain capacitance. e. Relationship between normal force and the value of floating-mountain capacitance. f. The floating-mountain capacitance exhibits insensitivity to tangential force during normal force measurements. g. Simulation of the magnetic field distribution of a coaxial circular Halbach magnetic film. h. Actual magnetic field distribution of a coaxial circular Halbach magnetic film along the radial direction. i. Relationship between tangential force and magnetic field strength under a fixed normal pressing distance of 2mm. j. The F³T sensing unit exhibits high sensitivity and accuracy in detecting the direction of tangential forces. k. The detection of the direction of tangential forces is independent of the magnitude. l. Temperature drift phenomenon affecting the magnetic sensing chip.

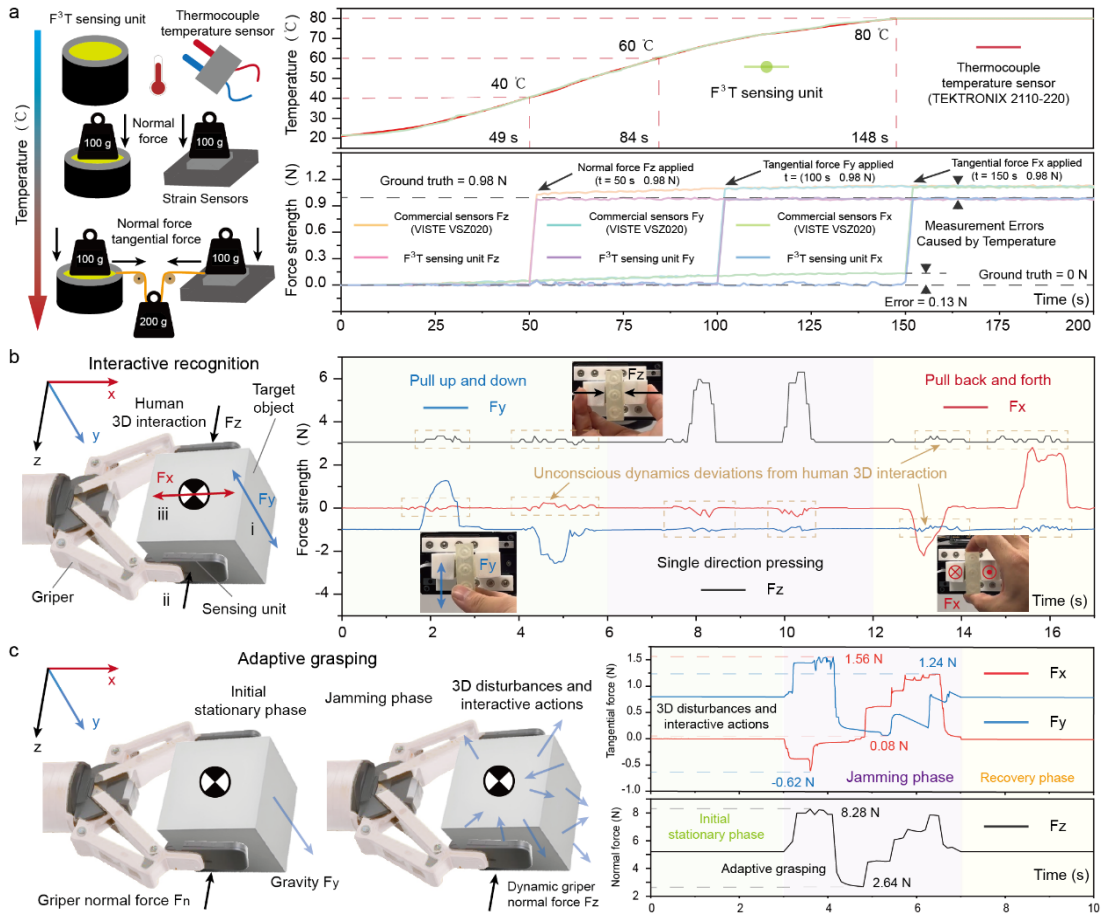

Figure 4. Performance of the F$^3$T sensing unit under static and dynamic conditions. a. Under static load and varying temperature environments, the temperature measurement of F$^3$T is comparable to that of a commercial thermocouple, and its ability to decouple multi-dimensional forces is superior to that of a traditional strain gauge force sensor. b. A robotic gripper integrated with the F$^3$T sensing unit can recognize human interaction movements. c. The robotic gripper with tactile feedback can achieve a higher level of adaptive and stable grasping under random external force disturbances.

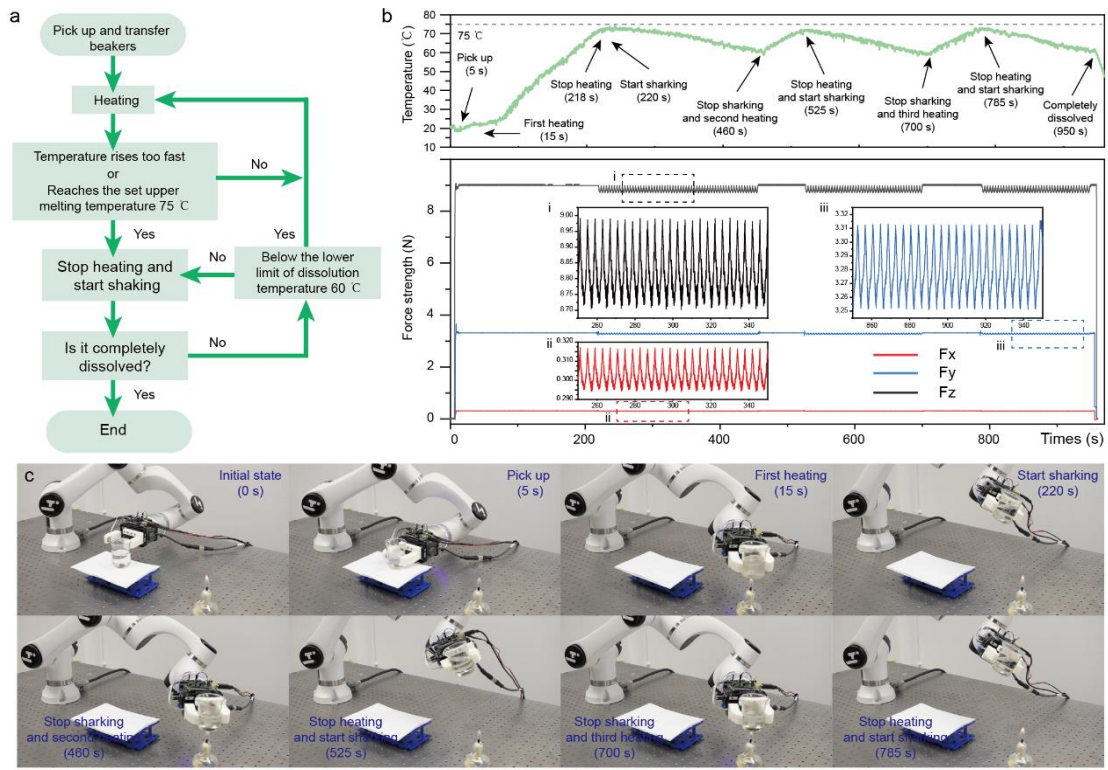

Figure 5. Demonstration of F³T for automated chemical reaction procedure. a. The process of automated preparation of PVA solution in laboratory. b. Detection of temperature and multi-dimensional forces by the F³T sensing unit during the preparation of the PVA solution. c. Optical images of key points during the demonstration.

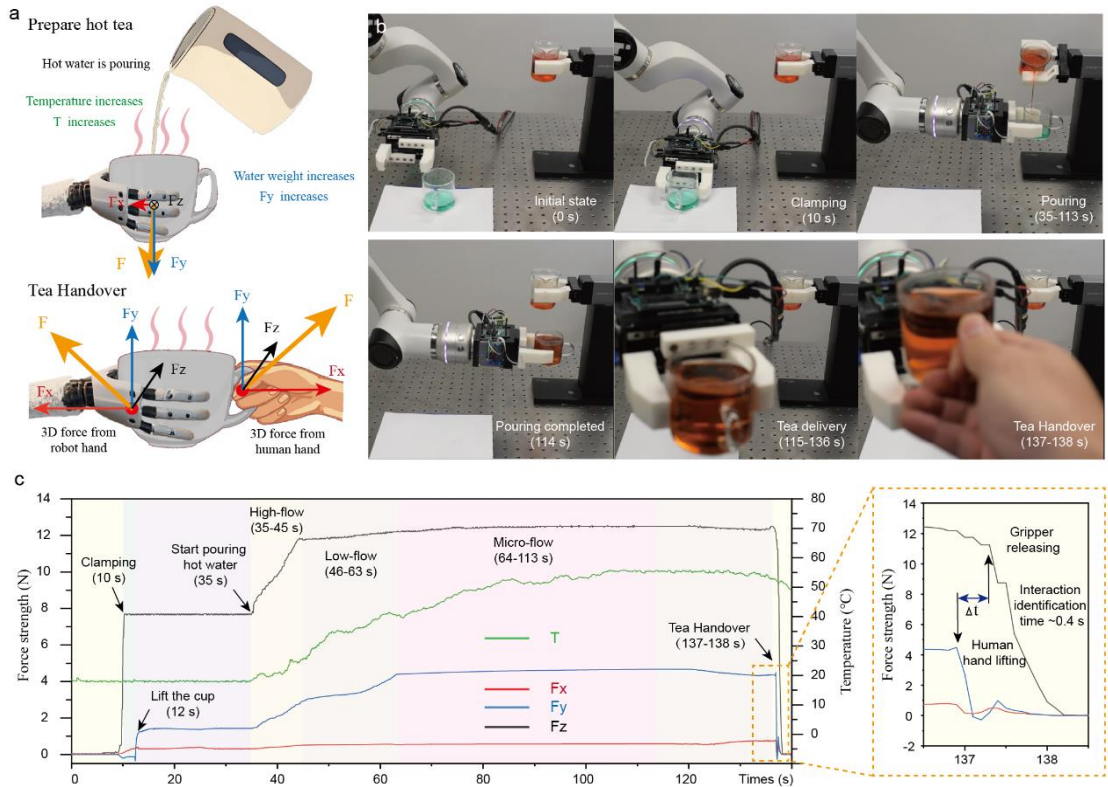

Figure 6. Demonstration of F³T for effective human-robot cooperation. a. Diagram showing changes in temperature and multi-dimensional forces detected by the robotic hand integrated with the F³T sensing unit during the tea-making and handover process. b. Optical images of key points during the demonstration. c. The actual changes in temperature and multi-dimensional forces obtained from the F³T sensing unit throughout the demonstration, where the recognition and confirmation of the handover action in approximately 0.4 s.